\documentclass[cameraready]{Interspeech}

\usepackage{newtxtext}
\usepackage{amsmath}
\usepackage{amssymb}
\usepackage{ifthen}

\usepackage{tabularx}
\definecolor{cvprblue}{rgb}{0.21,0.49,0.74}
\usepackage{pifont}
\newcommand{\cmark}{\ding{51}}
\newcommand{\xmark}{\ding{55}}

\title{Revisiting Active Speaker Detection: An In-the-Wild Benchmark for Generalization and Robustness}

\author[affiliation={1}, orcid=0009-0005-5922-8987, equalcontribution]{Le Thien Phuc}{Nguyen}
\author[affiliation={1}, equalcontribution]{Zhuoran}{Yu}
\author[affiliation={1}]{Khoa Quang Nhat}{Cao}
\author[affiliation={1}]{Yuwei}{Guo}
\author[affiliation={1}]{Tu Ho Manh}{Pham}
\author[affiliation={1}]{Tuan Tai}{Nguyen}
\author[affiliation={1}]{Toan Ngo Duc}{Vo}
\author[affiliation={2}]{Lucas}{Poon}
\author[affiliation={3}]{Tuan Khai}{Nguyen}
\author[affiliation={4}, orcid=0000-0002-2975-2519]{Soochahn}{Lee}
\author[affiliation={1}, correspondingauthor]{Yong Jae}{Lee}

\address{
    $^1$ University of Wisconsin - Madison, United States, $^2$ Oregon State University, United States \\
    $^3$ University of Sydney, Australia, $^4$ Kookmin University, Korea \\
    \url{https://github.com/plnguyen2908/UniTalk-ASD-code}.
}

\email{plnguyen6@wisc.edu, zhuoran.yu@wisc.edu, yongjaelee@cs.wisc.edu}


\begin{document}

\maketitle
\newcommand{\datasetname}{\textsc{UniTalk}}

\begin{abstract}

We present \datasetname{}, a novel dataset emphasizing challenging scenarios to enhance model generalization for the task of active speaker detection (ASD). 
Previously established benchmarks such as AVA predominantly comprise old movies and thus exhibit significant domain gaps with real-world video.
In contrast, \datasetname{} covers diverse video types reflecting challenging real-world conditions, including underrepresented languages, noisy backgrounds, and crowded scenes, while being on par with AVA in scale.
Extensive evaluations reveal that ASD remains unsolved under realistic conditions: state-of-the-art models near-perfect on AVA fail to reach saturation on \datasetname{}.
Conversely, models trained on \datasetname{} generalize better to modern in-the-wild datasets including Talkies and ASW. 
\datasetname{} thus establishes a new benchmark for ASD, providing researchers with a valuable resource for developing and evaluating versatile and resilient models. 
\end{abstract}
\section{Introduction}
\label{sec:intro}

Active speaker detection (ASD)~\cite{alcazar2021maas, ADENet, tao2021someone, min2022learning, liao2023light, wang2024loconet, jung2024talknce} aims to identify whether a visible person in a video is  speaking. This task plays a critical role in various downstream applications, including speaker diarization~\cite{lin2019lstm, wang2018speaker}, audiovisual speech recognition~\cite{wang2022multi, afouras2018deep, ma2021end}, and human-robot interaction~\cite{kang2023video, sheridan2016human, skantze2021turn}. To support the development of ASD models, several benchmark datasets have been proposed~\cite{roth2020ava, kim2021look, alcazar2021maas}, most notably the AVA-ActiveSpeaker dataset~\cite{roth2020ava}, which is constructed \textit{entirely from movie content}. AVA-ActiveSpeaker has become the de facto benchmark for evaluating ASD models and has driven significant progress in the field, with recent methods reporting nearly perfect mAP scores (e.g., $>95\%$~\cite{wang2024loconet, jung2024talknce}), leading many to consider ASD a solved problem in practice.


While AVA-ActiveSpeaker~\cite{roth2020ava} has been instrumental in driving progress, its reliance on movie data limits its ability to represent the complexities of real-world scenarios. In practice, ASD models must handle a wide range of challenges that are less common or absent in movie content, such as underrepresented spoken languages, noisy backgrounds (e.g., street sounds, music, or overlapping speech), and crowded scenes involving multiple people, occlusions, or dynamic camera motion. These factors are critical for deployment in settings like video conferencing, social media, and live broadcasts. However, the lack of benchmark coverage along these axes makes it difficult to assess model robustness or make meaningful improvements in generalization.

\begin{figure*}[t]
  \centering
  \includegraphics[width=\linewidth]{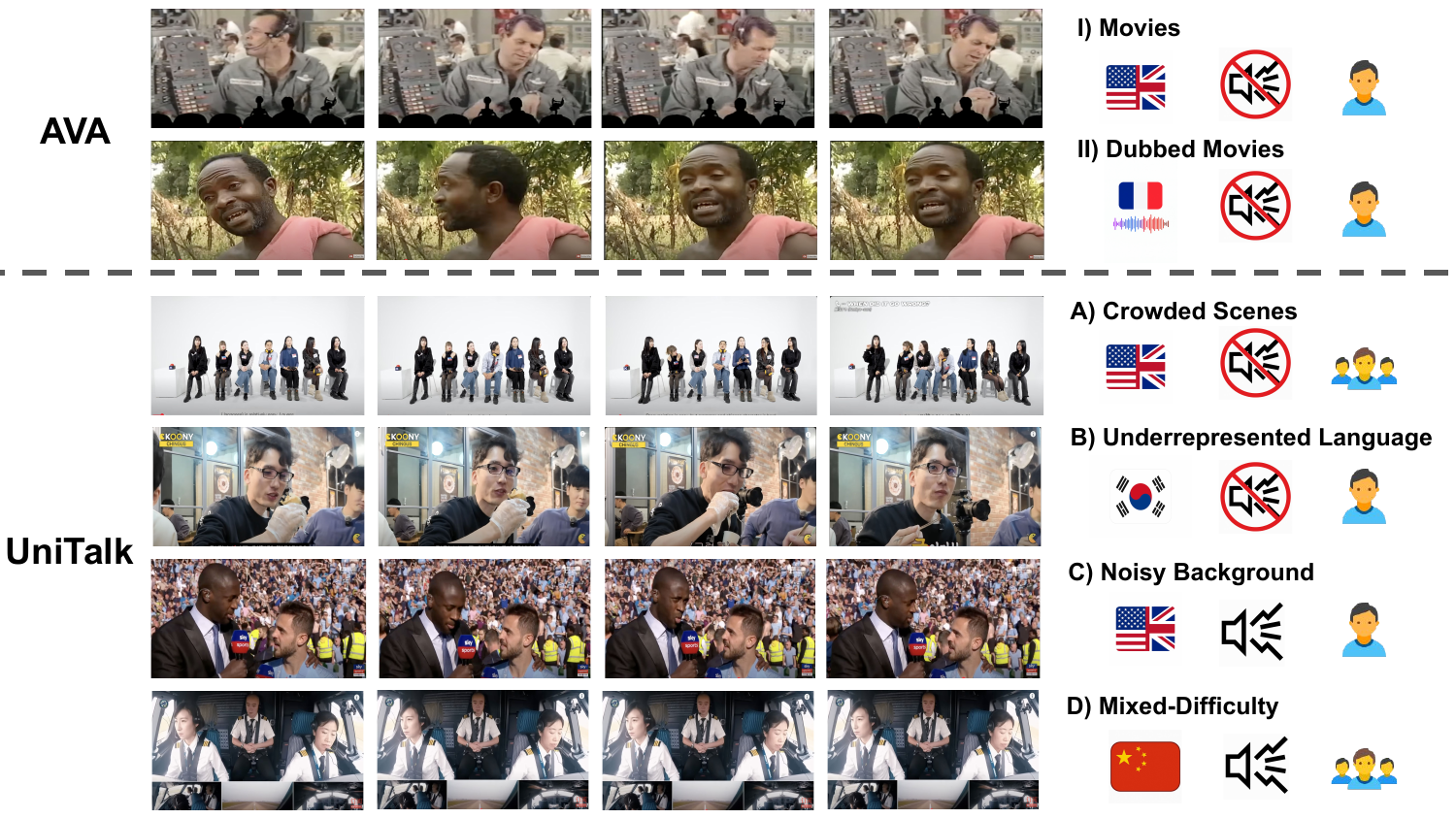}



\caption{\textbf{Comparison between AVA and \datasetname{}.} AVA~\cite{roth2020ava} primarily consists of movie content often with clean audio and simple visual composition. It also includes dubbed videos, where the audio is artificially overlaid and may not align with visible speech, potentially limiting the reliability of audiovisual supervision. In contrast, \datasetname{} features diverse real-world scenarios, including crowded scenes, underrepresented languages, noisy backgrounds, and combinations thereof. Each row shows a representative clip from a subcategory in \datasetname{}, with icons indicating language, noise level, and visual complexity.}
\label{fig:viz}
\end{figure*}

To address this gap, we introduce \datasetname{}, a new benchmark dataset for active speaker detection in real-world scenarios. While some prior datasets include online videos~\cite{alcazar2021maas, kim2021look}, they are neither curated to reflect the key challenges of real-world deployment nor comparable to AVA~\cite{roth2020ava} in scale. In contrast, \datasetname{} is constructed with an emphasis on diversity across multiple axes of difficulty, including underrepresented languages, noisy backgrounds, and crowded scenes, such as multiple visible speakers speaking concurrently or in overlapping turns. The dataset contains over 44.5 hours of video with frame-level active speaker annotations across 48,693 speaking identities, spanning a broad range of video types that reflect real-world conditions across these targeted difficulty axes. Fig.~\ref{fig:viz} highlights the key advantages of \datasetname{} over AVA.

\textbf{Evaluation}. To provide a standardized metric for comparison with existing benchmarks and to highlight the greater difficulty and headroom for improvement in \datasetname{}, we first follow prior work~\cite{roth2020ava} and evaluate over the full test set (i.e., for each visible person in each video frame, we evaluate whether the model correctly predicts active speaking status). In addition, unlike prior benchmarks, \datasetname{} enables fine-grained evaluation through a set of curated subcategories, each designed to stress a specific axis of difficulty. Specifically, the test set is further partitioned into four subsets: (1) \textit{underrepresented languages}, consisting of videos in languages that are less prevalent than English in both existing benchmarks and online media (e.g., East Asian languages), paired with clean audio and simple scenes; (2) \textit{noisy backgrounds}, containing videos with strong ambient noise but in well-represented languages such as English; (3) \textit{crowded scenes}, featuring visually challenging conditions such as multiple visible speakers or frequent occlusions; and (4) \textit{hard examples from mixed-difficulty}, which contain at least two of the above difficulty factors. This protocol supports more detailed analysis and highlights failure modes not evident from overall scores.


\textbf{Key Findings.} 
Although state-of-the-art ASD models report over 95 mAP on AVA~\cite{roth2020ava}, their performance drops substantially on \datasetname{}, with the strongest method achieving only 83.2 mAP. Performance further degrades under compounded real-world challenges (77.9 mAP on the Hard subset), revealing significant headroom under realistic conditions. This gap is not due to annotation noise or unrealistic difficulty. Models trained on \datasetname{} generalize consistently across AVA, Talkies, and ASW, achieving 88.0, 91.4, and 90.4 mAP respectively, while models trained on prior benchmarks exhibit strong in-domain performance but poor cross-dataset transfer. 

Moreover, pretraining on \datasetname{} enables rapid adaptation: a model fine-tuned with only 3 hours of AVA data reaches 92.4 mAP, approaching full-data performance. 
Together, these findings suggest that existing benchmarks may overestimate real-world readiness, and that \datasetname{} provides a more realistic and transferable foundation for ASD evaluation and training.

\section{Related Work}

\textbf{Active Speaker Detection Datasets.}
Early audiovisual speaker detection datasets such as VoxCeleb~\cite{nagrani2017voxceleb} and the Columbia dataset~\cite{chakravarty2016cross} focus on relatively constrained scenarios including interviews and monologue-style speech. 
AVA-ActiveSpeaker~\cite{roth2020ava} later became the most widely used benchmark with large-scale frame-level annotations, but its heavy reliance on movie content limits coverage of real-world conditions. 
More recent datasets, including Talkies~\cite{alcazar2021maas}, ASW~\cite{kim2021look}, and VoxConverse~\cite{Chung_2020}, incorporate web videos and offer increased diversity, yet remain limited in scale and do not systematically capture practical deployment challenges. 
To address these limitations, we introduce \datasetname{}, a comprehensive benchmark designed to evaluate model robustness across underrepresented languages, diverse noise conditions, and crowded visual environments.

\textbf{Active Speaker Detection Methods.}
Active speaker detection (ASD) aims to determine whether a visible person in a video frame is speaking, requiring effective audiovisual modeling. 
Existing approaches generally adopt either multi-stage training, where feature encoders and context models are optimized separately~\cite{UniCon, alcazar2020active, kopuklu2021design, min2022learning}, or single-stage frameworks that jointly learn these components~\cite{tao2021someone, liao2023light, wang2024loconet, jung2024talknce}. 
Recent work emphasizes long-term context modeling using recurrent networks, attention mechanisms, or hybrid architectures~\cite{liao2023light, wang2024loconet, datta2022asd}. 
Despite strong performance on benchmarks such as AVA, existing methods still exhibit notable performance gaps under more challenging real-world conditions, highlighting the need for more diverse and realistic datasets.
\section{\datasetname{} Dataset}

\subsection{Preliminary: Active Speaker Detection}

\begin{figure*}[t]
  \centering
    \includegraphics[width=\linewidth]{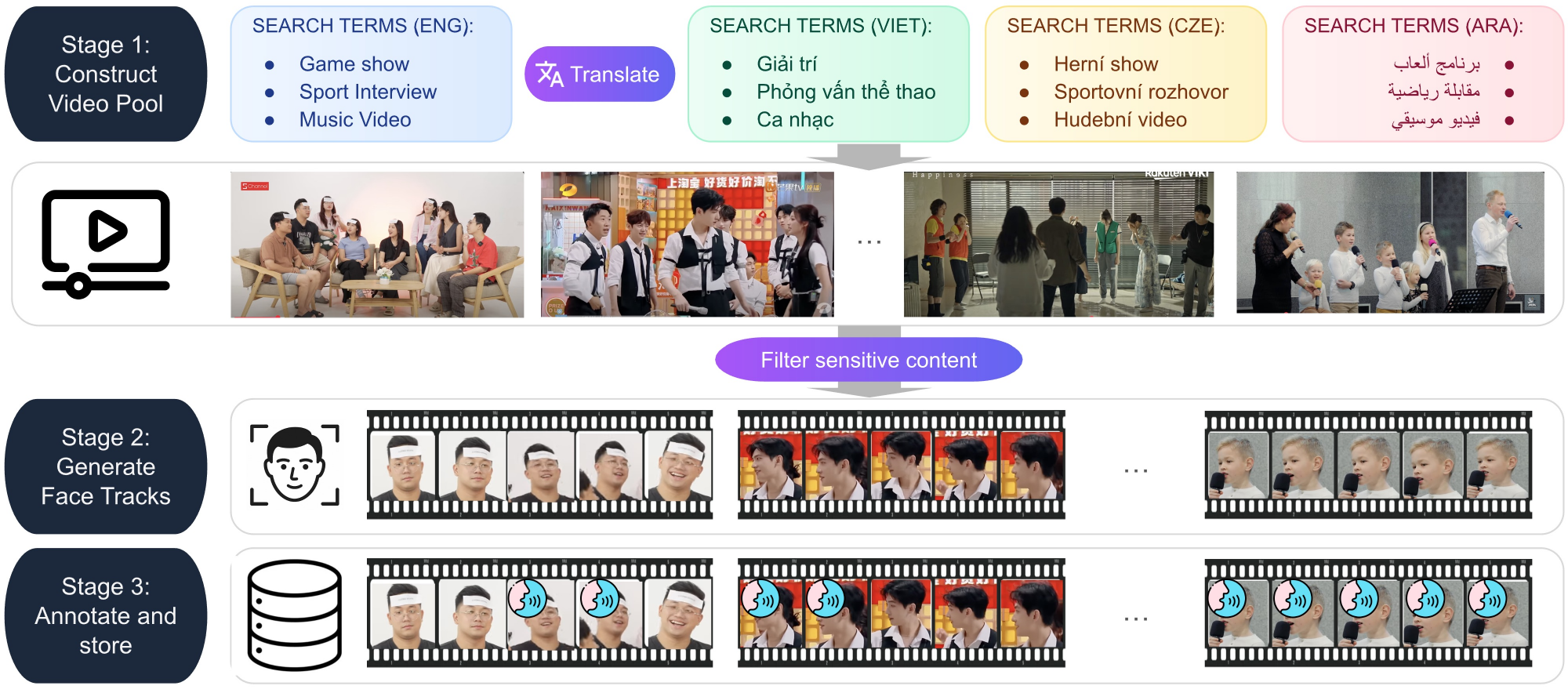}
  \caption{\textbf{Data curation pipeline.} Our data curation pipeline consists of four distinct stages: (1) video sourcing to construct an initial pool of candidate clips, (2) content filtering to remove videos containing sensitive or inappropriate material, (3) face track generation to convert raw videos into structured face sequences, and (4) annotation and storage for benchmark use.}
  \label{fig:pipeline}
\end{figure*}   

The goal of active speaker detection (ASD) is to make a binary decision
$Y \in [0,1]^T$ given a face track $V \in \mathbb{R}^{T \times H \times W}$
and an audio track $A \in \mathbb{R}^{4T \times M}$, where $T$ is the temporal
length of the face track, $H$ and $W$ are the spatial dimensions of each face,
and $M$ is the number of Mel-spectrogram frequency bins. The $4T$ factor
reflects the sampling-rate mismatch between modalities: for a clip of $\tau$
seconds, STFT with a $10$\,ms hop and $25$\,ms window yields
$1 + (\tau - 0.025)/0.01 \approx 100\tau$ audio frames, while video sampled at
$25$\,fps yields $25\tau$ frames. Hence, we have $4T$ audio frames per $T$ video frames,
consistent with prior benchmarks~\cite{roth2020ava}.

State-of-the-art ASD models~\cite{alcazar2021maas, kopuklu2021design,
tao2021someone, min2022learning, liao2023light, wang2024loconet,
jung2024talknce} comprise an audio-visual encoder and a context modeling module.
The visual and audio encoders $\mathcal{F}_v, \mathcal{F}_a$ map $V$ and $A$ to
features $f_v \in \mathbb{R}^{T \times D_v}$ and
$f_a \in \mathbb{R}^{T \times D_a}$, which are concatenated along the embedding
dimension into $f_{av} \in \mathbb{R}^{T \times (D_v + D_a)}$. A context module
$\mathcal{C}$ then produces the context-aware feature $f'_{av}$ for the main
prediction. Three linear classifiers operate on $f'_{av}$, $f_a$, and $f_v$,
trained jointly with
\begin{equation}
\mathcal{L}_{asd} = \lambda_{av} \mathcal{L}_{av} + \lambda_{a} \mathcal{L}_{a}
+ \lambda_v \mathcal{L}_v,
\end{equation}
where each term is the cross-entropy between $Y$ and the prediction from
$f'_{av}$, $f_a$, and $f_v$, respectively; $\mathcal{L}_a$ and $\mathcal{L}_v$
act as auxiliary losses encouraging attention to both modalities.
\subsection{Data Curation}
\label{sec:data_curation}

Our IRB-approved data curation process is designed to construct a large-scale benchmark for active speaker detection that reflects the diversity and complexity of real-world audiovisual conditions. Specifically, the pipeline targets coverage across three critical axes of difficulty: underrepresented languages, noisy backgrounds, and crowded scenes with multiple visible speakers. While not every video includes all these challenges simultaneously, the dataset as a whole is curated to provide meaningful representation along each axis. The full pipeline, depicted in Figure~\ref{fig:pipeline}, consists of three stages (candidate video sourcing, face track generation, and annotation) designed to ensure both scale and annotation quality while preserving the diversity necessary for evaluating model robustness.

\begin{table*}[t]
  \centering
  \setlength{\tabcolsep}{6pt}
  \footnotesize
  
  \caption{\textbf{Quantitative comparison between ASD datasets.} All statistics are computed over the combined training and test sets of each dataset. The highest value in each column is shown in \textbf{bold}.}
  \vspace{-3mm}
  \label{data:cross_comparison}
  \resizebox{\linewidth}{!}{
  \begin{tabular}{l|ccccccccc}
    \toprule
    Dataset & Year & \shortstack{Total\\hours} & \shortstack{Total\\face tracks} & \shortstack{Total\\face crops} & 
    \shortstack{Avg speakers\\per frame} & \shortstack{Multi-source\\domain} & In-the-wild & Multilingual & Axes Evaluation \\
    \midrule
    AVA~\cite{roth2020ava} & 2020 & 38.5  & 37,738 & 3.4M   & 1.5 & \xmark & \xmark & \cmark & \xmark\\
    ASW~\cite{kim2021look} & 2021 & 23    & 8,000  & 407K   & 1.9 & \xmark & \cmark & \xmark & \xmark \\
    Talkies~\cite{alcazar2021maas} & 2021 & 4.2   & 23,508 & 799K  & 2.3 & \cmark & \cmark & \xmark & \xmark \\
    \datasetname{} & 2025 & \textbf{44.5} & \textbf{48,693} & \textbf{4M} & \textbf{2.6} & \cmark & \cmark & \cmark & \cmark \\
    \bottomrule
  \end{tabular}
  }
\end{table*}

\begin{figure*}[t]
  \centering
  \resizebox{0.9\textwidth}{!}{
    \begin{minipage}{0.5\textwidth}
    \centering
    \includegraphics[width=\linewidth]{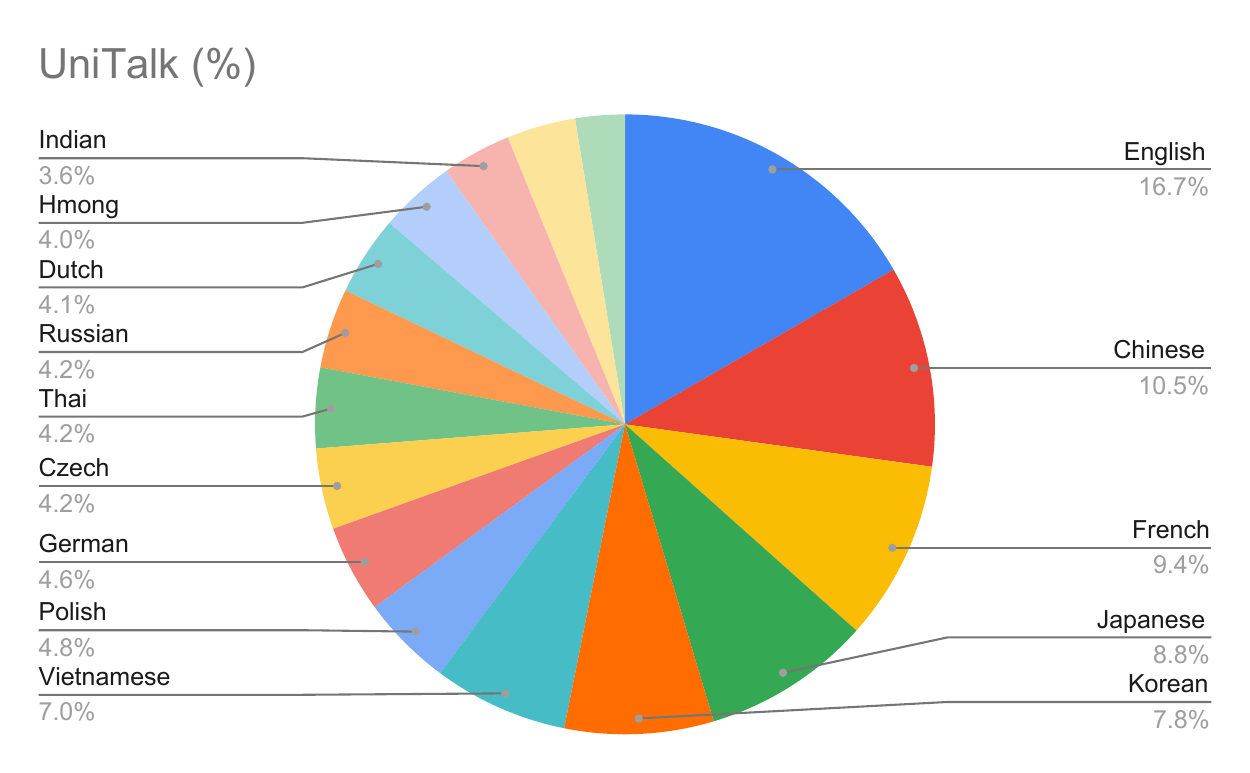}
  \end{minipage}
  \hfill
  \begin{minipage}{0.5\textwidth}
    \centering
    \includegraphics[width=\linewidth]{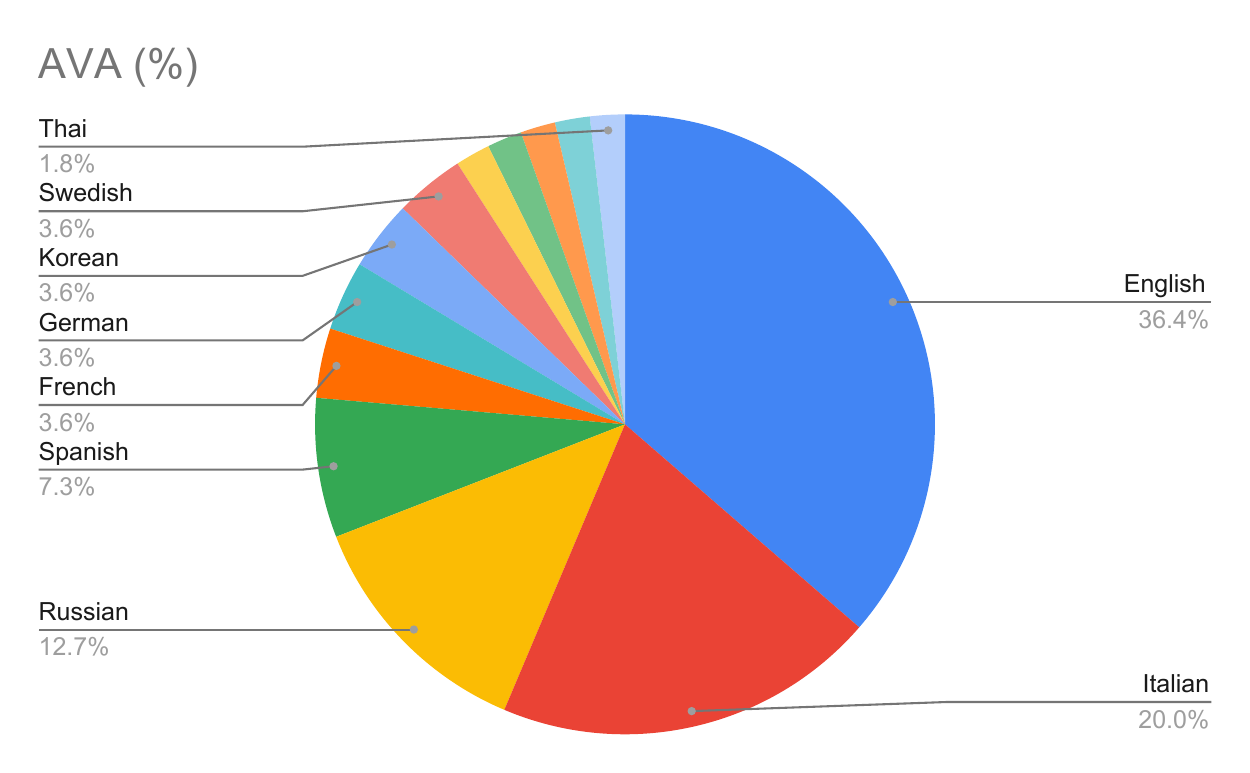}
  \end{minipage}
  \hfill
  }
  \caption{\textbf{Language distribution in \datasetname{} vs.~AVA.} \datasetname{} covers a wider range of languages, particularly with stronger representation of East Asian languages e.g., Chinese, Korean, and Japanese. In contrast, AVA primarily consists of Indo-European languages, limiting its linguistic diversity.}
  \label{fig:lang_dist}
\end{figure*}

\textbf{Candidate Video Sourcing.}
To construct a diverse pool of speaking scenarios, we first generate keyword search terms (with the help of GPT-4~\cite{achiam2023gpt}) corresponding to video scenes likely to exhibit either visual or acoustic complexity, such as multi-person talk shows, press conferences, classroom discussions, sports interviews, panel debates, etc. Following prior work~\cite{alcazar2021maas, kim2021look}, we use YouTube as the primary source of videos. The search keywords are then translated into multiple languages to encourage linguistic diversity and used to retrieve candidate videos across different regions. To ensure high annotation quality and reduce downstream noise, we apply a combination of automated and manual filtering: Videos are discarded if they exhibit excessive face occlusions, very low resolution (below 480 pixels on the shorter side), or poor audio conditios, such as missing speech, overpowering background music, or severe reverberation. We also exclude videos containing sensitive or inappropriate content to uphold ethical standards for data collection and annotation.

\textbf{Face Track Generation.}
To support dense, frame-level speaker annotation, we generate face tracks using an automatic face detection and tracking pipeline. Candidate faces are first detected using S3FD~\cite{Zhang_2017_ICCV} and then linked across frames using a greedy tracking algorithm based on spatial overlap and visual similarity. Tracks are smoothed using Gaussian kernel filtering on keypoint trajectories, and gaps shorter than 0.2 seconds are linearly interpolated to ensure temporal continuity. To ensure annotation quality and feasibility, we follow the filtering criteria established by AVA-ActiveSpeaker~\cite{roth2020ava}: we retain only face tracks that are at least 1 second in duration to provide sufficient temporal context. Each retained track is paired with synchronized audio and video playback to facilitate accurate speaker labeling. Occasional tracking failures, such as identity switches or false-positive detections, are manually flagged and discarded by annotators during the annotation stage. This step ensures that only high-quality face tracks are retained for final annotation. In total, the process yields 48,693 face tracks across 4.0 million faces, forming the basis for robust and scalable active speaker labeling in \datasetname{}.

\textbf{Annotation Protocol.}
The annotation phase consists of a two-stage, multi-pass labeling process to ensure both high recall and high precision. In the first stage, multiple annotators independently review each face track and label whether the person is actively speaking at each frame. To maximize recall, annotators are instructed to label any moment where a person appears to be producing a verbal signal. In the second stage, a different set of annotators verifies these initial labels, either confirming or correcting the speaking status. Final labels are retained only when a majority consensus among annotators is reached, reinforcing reliability and reducing subjective bias. We follow the annotation criteria established by AVA-ActiveSpeaker~\cite{roth2020ava} for determining what constitutes speaking. Specifically, a person is considered to be speaking if they are producing a verbal signal that carries semantic content: this includes normal speech, shouting, singing, or calling out. Non-verbal vocalizations such as coughing, laughing, sneezing, or other incidental mouth movements without semantic content are not labeled as speaking, even if the mouth is visibly active. More details on the annotation criteria can be found in Table~\ref{tab:speaking_vs_nonspeaking}. These guidelines aim to ensure consistent, high-quality annotations across diverse audiovisual conditions, including crowded and noisy scenes where cues may be ambiguous. In total, the dataset comprises 44.5 hours of densely annotated video. To support benchmark development, the data is partitioned at the video level into a training split (33.4 hours) and a test split (11.1 hours), ensuring that no speakers or conversational contexts are shared between splits. This prevents data leakage and supports rigorous evaluation of generalization.


\begin{table}[t]
\centering
\small
\setlength{\tabcolsep}{4pt}
\renewcommand{\arraystretch}{1.10}
\caption{Annotation criteria for actively speaking vs non-speaking.}
\vspace{-2mm}
\label{tab:speaking_vs_nonspeaking}

\begin{tabularx}{\columnwidth}{@{}X X@{}}
\toprule
\textbf{Speaking} & \textbf{Non-speaking} \\
\midrule

\begin{itemize}\setlength\itemsep{0pt}\setlength\parskip{0pt}\setlength\topsep{0pt}\setlength\partopsep{0pt}\setlength\leftmargin{1.2em}
  \item Conversational speech (full sentences or phrases).
  \item Short verbal responses (e.g., ``Yes,'' ``No,'' ``Go,'' ``Okay'').
  \item Speech in presentations, interviews, or dialogues.
  \item Vocal fillers (e.g., ``um,'' ``ah,'' ``hmm'').
  \item Audible mumbling with speech-like intent.
  \item Singing (with or without accompaniment).
\end{itemize}
&
\begin{itemize}\setlength\itemsep{0pt}\setlength\parskip{0pt}\setlength\topsep{0pt}\setlength\partopsep{0pt}\setlength\leftmargin{1.2em}
  \item Laughter, sighs, groans, grunts, coughing, humming.
  \item Breath sounds without speech articulation.
  \item Mouthing without producing sound.
  \item Non-verbal communication (e.g., nodding, waving).
  \item A/V desynchronization (e.g., off-screen narration).
\end{itemize}
\\

\bottomrule
\end{tabularx}

\vspace{-10mm}
\end{table}




\subsection{Dataset Statistics}

  

\begin{figure*}[ht]
  \centering
  \resizebox{1.0\textwidth}{!}{
    \begin{minipage}{0.32\textwidth}
    \centering
    \includegraphics[width=\linewidth]{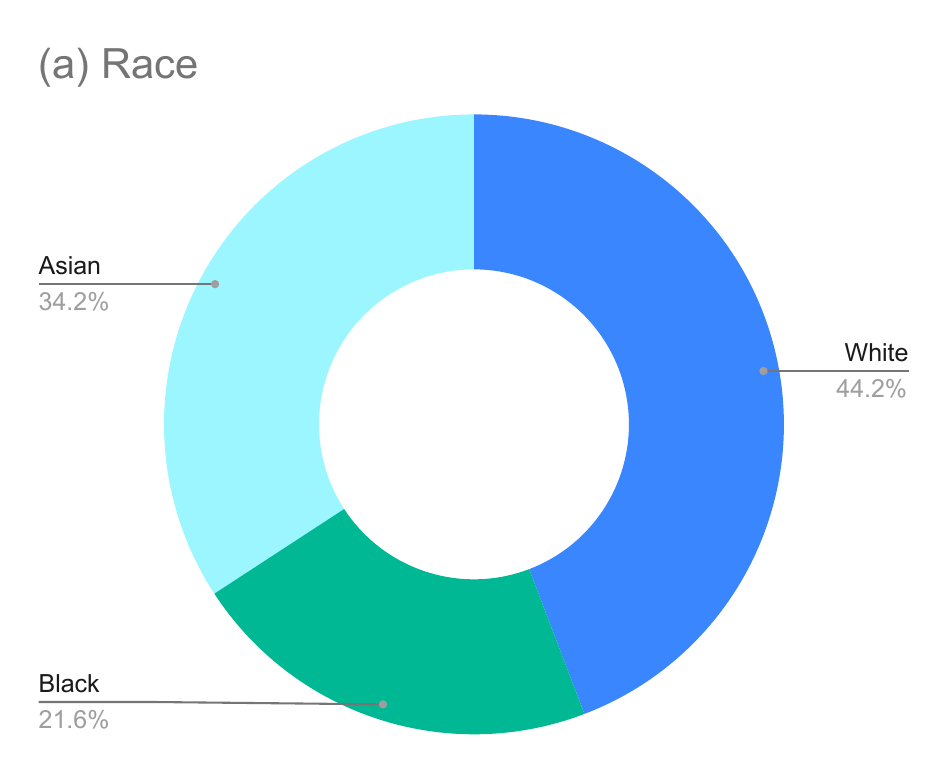}
  \end{minipage}
  \hfill
  \begin{minipage}{0.32\textwidth}
    \centering
    \includegraphics[width=\linewidth]{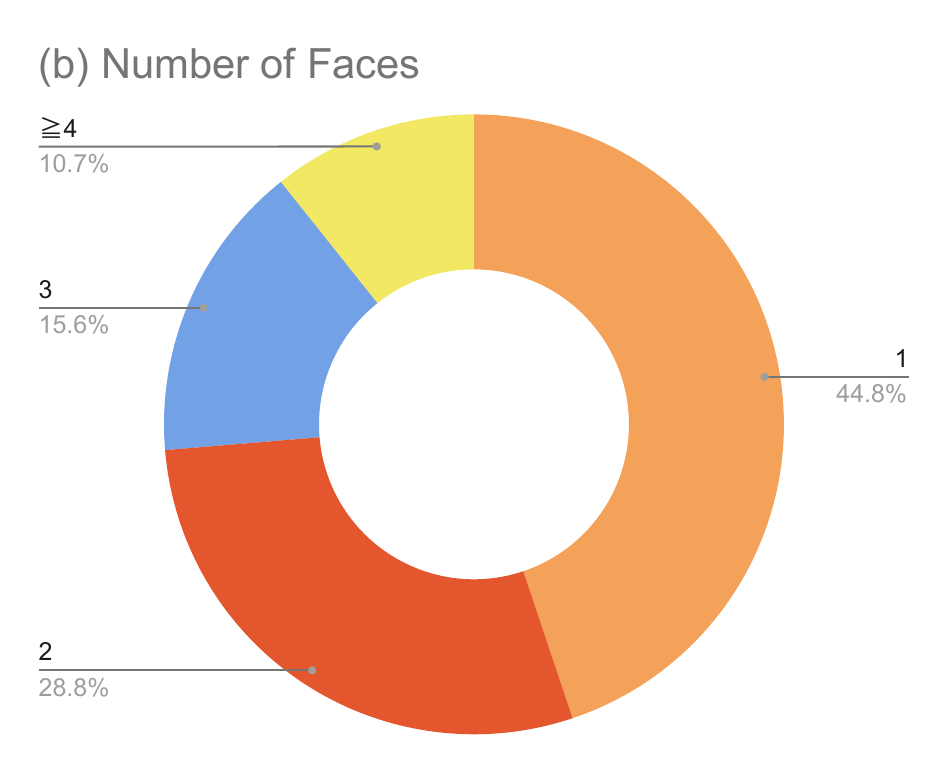}
  \end{minipage}
  \hfill
  \begin{minipage}{0.32\textwidth}
    \centering
    \includegraphics[width=\linewidth]{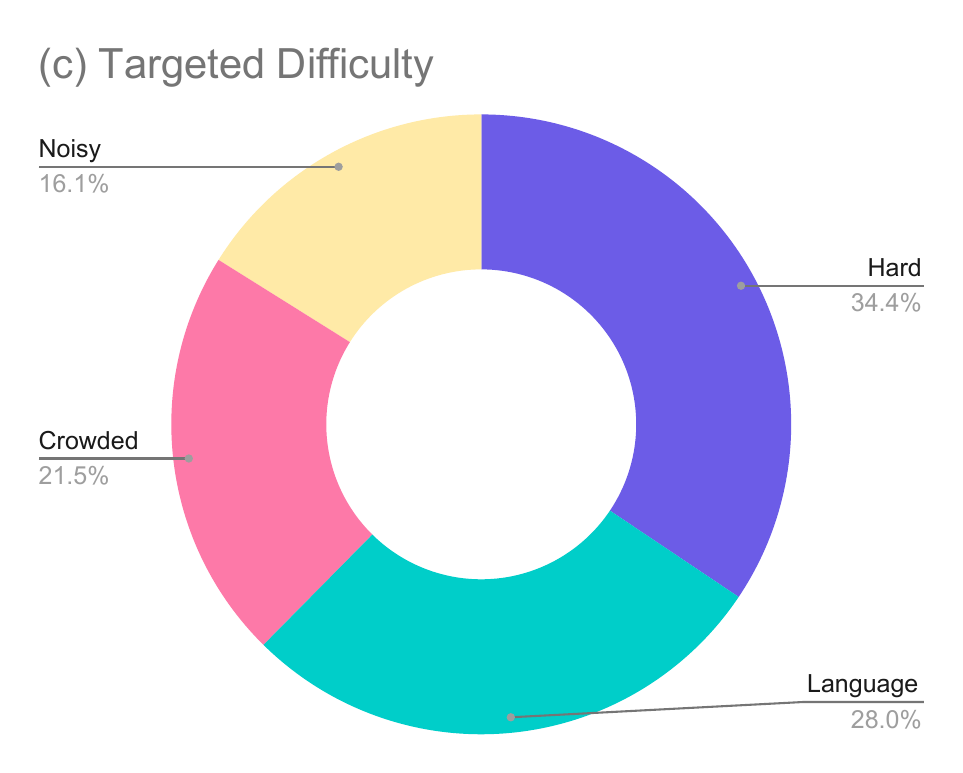}
  \end{minipage}
  }
    \caption{\textbf{Dataset Composition Overview.} 
    (a) Race distribution of visible speakers. 
    (b) Number of visible faces per frame, reflecting the range of visual complexity. 
    (c) Breakdown of test set according to targeted difficulty categories used for evaluation.} 
  \label{fig:demographics}
\end{figure*}

\textbf{Quantitative Comparison with Existing ASD Benchmarks.}
Table~\ref{data:cross_comparison} presents a quantitative comparison between \datasetname{} and several widely used benchmarks in active speaker detection, including AVA~\cite{roth2020ava}, ASW \cite{kim2021look}, and Talkies~\cite{alcazar2021maas}. \datasetname{} offers the largest scale, with 44.5 hours of annotated video, surpassing AVA (38.5 hours), ASW (23 hours), and Talkies (4.2 hours). It also provides the highest number of face tracks (48,693) and face crops (4 million), indicating greater coverage of speaker appearances. Furthermore, \datasetname{} exhibits the highest speaker density, averaging 2.6 visible speakers per frame compared to 2.3 for Talkies, 1.9 for ASW, and 1.5 for AVA, reflecting the increased interaction complexity in our benchmark.

\textbf{Demographic and Language Diversity.}
\datasetname{} features an ethnically diverse set of speaking identities, with 44.2\% White, 34.2\% Asian, and 21.6\% Black individuals, ensuring a broad demographic representation. In terms of language distribution (Figure~\ref{fig:lang_dist}), \datasetname{} also improves upon the linguistic coverage seen in AVA~\cite{roth2020ava}, which primarily contains Indo-European languages. While English remains the dominant language in both datasets, AVA underrepresents East Asian languages such as Chinese, Korean, and Japanese, languages that are substantially better represented in \datasetname{}. This improved linguistic balance enables more robust evaluation of ASD models in multilingual and cross-cultural scenarios.

\subsection{Benchmarking Task and Evaluation}

\datasetname{} is designed to serve as a standardized benchmark for active speaker detection under real-world conditions. Following prior work~\cite{roth2020ava}, we formulate ASD as a binary classification task at the face track level: for each video frame, the model must determine whether a given visible person is actively speaking. We adopt mean average precision (mAP) as the primary evaluation metric, following common benchmark practice~\cite{roth2020ava, alcazar2021maas, kim2021look}, enabling consistent and comparable evaluation across models. 

To facilitate deeper analysis of model robustness, we define a set of diagnostic
subgroups aligned with the core axes of difficulty in \datasetname{}: language
diversity, background noise, and visual crowding. For the first three, we
\emph{explicitly isolate} each difficulty factor, selecting examples that exhibit
the target condition while controlling for the others, so that model behavior can
be assessed under controlled stress conditions:
\begin{itemize}[leftmargin=*, nosep]
  \item \textbf{Underrepresented Languages:} scenes where the dominant language
  is non-English, with minimal background noise and visual complexity to isolate
  linguistic variation.
  \item \textbf{Noisy Backgrounds:} scenes with strong ambient noise or music
  where speech remains intelligible and the speaker is clearly visible; language
  and crowding are controlled to isolate acoustic difficulty.
  \item \textbf{Crowded Scenes:} scenes with multiple visible speakers,
  occlusions, or rapid camera motion, with language and background noise
  controlled to isolate visual complexity.
  \item \textbf{Hard Examples:} scenes combining at least two difficulty axes,
  such as overlapping speakers in underrepresented languages or noisy
  conversations in visually complex settings.
\end{itemize}

To construct these subgroups and quantify the two key sources of real-world
complexity---scene crowdedness and background noise---we compute two diagnostic
metrics:
\begin{itemize}[leftmargin=*, nosep]
\item \textbf{Visual complexity.} The average number of visible faces per frame,
obtained via face detection. A threshold of two faces serves as a mid-level
cutoff, informed by AVA statistics~\cite{wang2024loconet} where 99\% of samples
fall below it, separating simple from crowded scenes.
\item \textbf{Background noise level.} The Root Mean Square (RMS) energy of the
background audio, computed after removing speech segments with a Voice Activity
Detection (VAD) tool~\cite{SileroVAD}. A threshold of 0.03 RMS distinguishes low
from high noise levels.
\end{itemize}

\begin{figure}[ht]
  \centering
    \includegraphics[width=\linewidth]{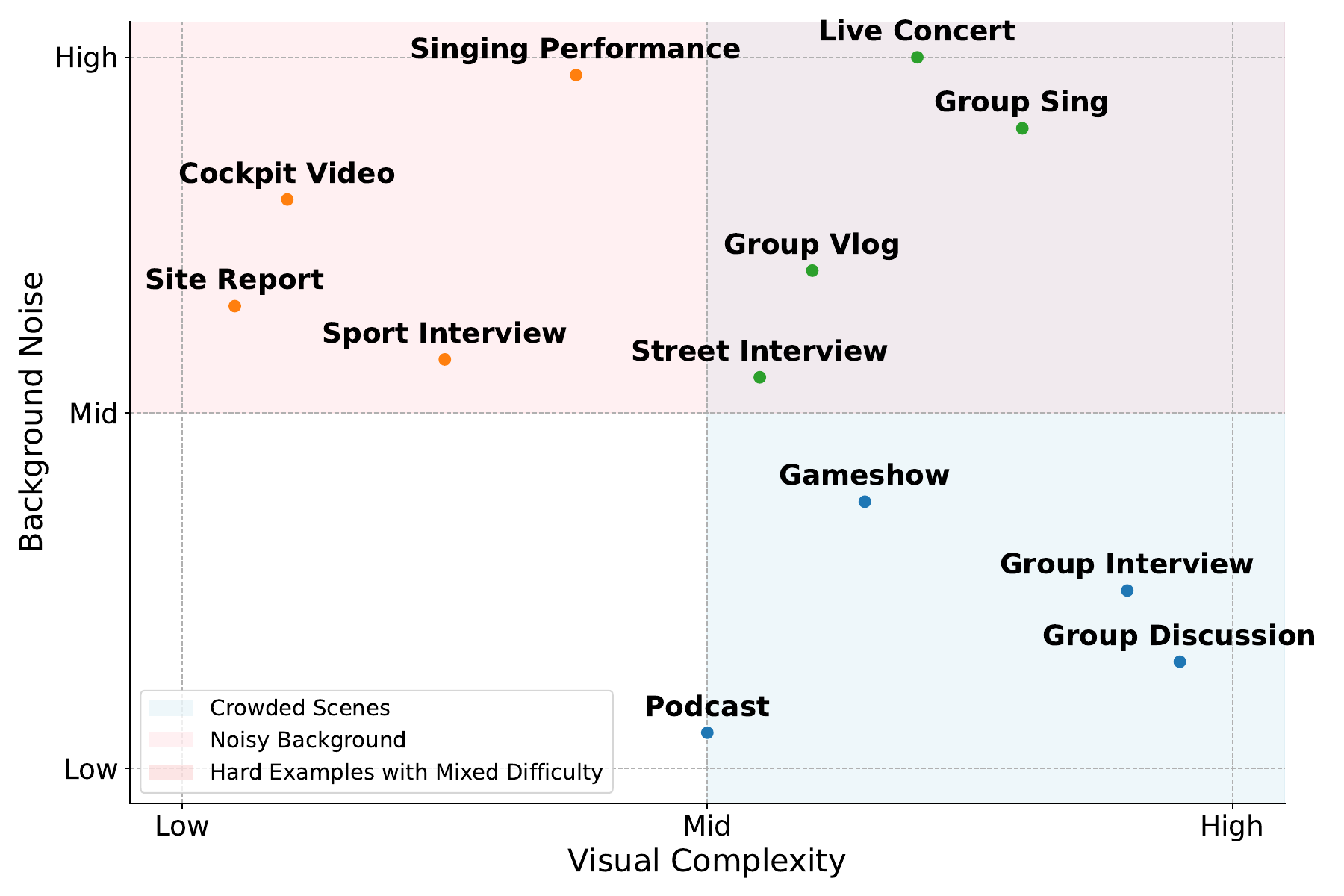}
  \caption{\textbf{Difficulty space of video search terms.} Each point
  represents a video category, plotted by the average number of faces per frame
  (x-axis, visual complexity) and average background noise level (y-axis,
  measured via RMS after VAD).}
  \label{fig:prompt}
  \vspace{-25pt}
\end{figure}

We plot each video search term in this 2D space (Figure~\ref{fig:prompt}), which
provides an interpretable overview of the diversity in the collected data. We highlight three shaded regions corresponding to
the different axes of difficulty: \textbf{crowded scenes} (bottom right), \textbf{noisy backgrounds} (top left), and \textbf{hard examples} (top right). As shown in Figure~\ref{fig:demographics}c, the
test set contains substantial coverage across all axes: 28.0\% of samples feature
underrepresented languages, 21.5\% involve visually crowded scenes, 16.1\%
contain noisy audio, and 34.4\% fall into the hard example category.
Figure~\ref{fig:demographics}b further highlights that over 55\% of frames
contain two or more visible faces, reinforcing the need for robust modeling in
multi-speaker scenarios.

\subsection{Data Use and Ethics}
The study protocol was reviewed by the institutional review board (IRB), classified as minimal risk, and determined not to constitute human-subject research under DHHS and FDA regulations. For anonymity, the approval document will be included in the supplementary upon acceptance.

This dataset is designed to advance robust, generalizable active speaker detection (ASD), with potential benefits for accessibility, video understanding, and human--computer interaction through greater real-world diversity. As with similar datasets, misuse (e.g., surveillance or privacy-violating applications) is possible and strictly prohibited. The dataset will be released under CC BY-NC 4.0, which forbids facial recognition, surveillance, and biometric identification uses.

Each entry contains only face tracks, audio, and human-written annotations following AVA's convention~\cite{roth2020ava}; no video content is redistributed. If an identifiable individual requests removal, the relevant segments and metadata will be removed from future releases.

\section{Experiments}

We conduct a series of experiments to assess the effectiveness of \datasetname{} as a challenging and representative benchmark for active speaker detection (ASD). First, we evaluate a range of state-of-the-art ASD models on \datasetname{} to understand their performance under realistic visual and acoustic conditions. Next, we compare major ASD datasets, AVA, Talkies, ASW, and \datasetname{}, by training and evaluating models across all combinations, highlighting differences in generalization and domain coverage. Finally, we demonstrate the utility of \datasetname{} as a training source by fine-tuning a model pretrained on \datasetname{} for the AVA benchmark, showing rapid adaptation and strong downstream performance. Together, these results position \datasetname{} as a valuable benchmark for both model evaluation and pretraining in real-world ASD scenarios.

\subsection{Training ASD Models on \datasetname{}}

We benchmark a range of representative active speaker detection (ASD) models on
\datasetname{}, including multi-stage architectures such as
ASDNet~\cite{kopuklu2021design} and ASC~\cite{alcazar2020active}, and end-to-end
and contrastive learning approaches like TalkNet~\cite{tao2021someone},
LoCoNet~\cite{wang2024loconet}, LASER~\cite{nguyen2025laser}, and
TalkNCE~\cite{jung2024talknce}. All models are trained from scratch on the
\datasetname{} training split and evaluated on its held-out test set using mean
average precision (mAP), which we compute over positive face detections by
ranking their predicted scores and measuring the area under the
precision-recall curve.

\begin{table*}[t]
  \centering
  \footnotesize
   \caption{\textbf{Performance of ASD models trained and evaluated on \datasetname{}.} Models are chosen to showcase different approaches, ranging from multi-stage systems to contrastive learning approaches. Results are reported in mAP. The highest score is shown in \textbf{bold}.}

  \begin{tabular}{cccccccc}
    \toprule
    & & & \multicolumn{5}{c}{\datasetname{}} \\
     \cmidrule(r){4-8}
    Model & Architecture & Train Data &Overall & Language &  Crowded & Noise & Hard\\
    \midrule
    ASDNet~\cite{kopuklu2021design}         & ResNeXt/BGRU    & \datasetname{}  & 20.6 & 30.8 & 17.5 & 14.8 &20.3 \\
    ASC~\cite{alcazar2020active}            & ResNet/LSTM    &  \datasetname{}  & 61.4 & 74.7 & 62.9& 53.4 & 57.3\\
    TalkNet~\cite{tao2021someone}           & ResNet/LIM     &  \datasetname{}  & 75.7 & 80.1 & 77.6 & 67.1 & 70.3\\
    LoCoNet~\cite{wang2024loconet}          & TalkNet/SIM     &  \datasetname{} & 82.2 & 85.8 & 84.6 & 80.0 & 76.2 \\
    LASER~\cite{nguyen2025laser} & LoCoNet/Lip Landmark & \datasetname{} & 82.2 & \textbf{86.7} & 83.7 & 81.6 & 75.8 \\
    TalkNCE~\cite{jung2024talknce}          & LoCoNet/NCE loss &  \datasetname{} & \textbf{83.2} & \textbf{86.7} & \textbf{84.9} & \textbf{84.1} & \textbf{77.9} \\
    \midrule
    TalkNCE~\cite{jung2024talknce}          & LoCoNet/NCE loss&  AVA~\cite{roth2020ava} & 77.5 & 84.9   & 81.0  & 80.1 & 64.8\\
    \bottomrule
  \end{tabular}
 
  \label{tab:models_on_our}
\end{table*}

\subsection{Implementation Details}


We implement two multi-stage ASD models~\cite{kopuklu2021design, alcazar2020active} and four single-stage ASD models~\cite{tao2021someone,nguyen2025laser, wang2024loconet,jung2024talknce}. For fair comparison, we follow each model's original setup. For LoCoNet~\cite{wang2024loconet}, LASER~\cite{nguyen2025laser} and TalkNCE~\cite{jung2024talknce}, we use a batch size of 4 and sample 200 frames per training example. Each model is trained with 25 epochs. For TalkNet~\cite{tao2021someone}, we use a batch size that contains at most 5000 frames, and the model is trained for 25 epochs. For single-stage ASD training objectives, we follow the LoCoNet and TalkNet, setting $\lambda_{av} = 1$, $\lambda_a = 0.4$, and $\lambda_v = 0.4$. For TalkNCE loss~\cite{jung2024talknce}, we set its weight to 0.3 as mentioned in the paper. Random resizing, cropping, horizontal flipping, and rotations are used as visual data augmentation operations and a randomly selected audio signal from the training set is added as background noise to the target audio~\cite{tao2021someone}. For the multi-stage training, we train ASC's encoder for 100 epochs before training its context module for 15 epochs \cite{alcazar2020active}. Similarly, we train ASDNet's encoder for 70 epochs and its context module for 10 epochs \cite{kopuklu2021design}. For both ASC~\cite{alcazar2020active} and ASDNet~\cite{kopuklu2021design}, we set $\lambda_{av} = \lambda_{a} = \lambda_{v} = 1$ in the first stage and $\lambda_{av}=1$ in the second stage. Finally, we use Adam~\cite{diederik2014adam} as our optimizer.

\subsection{Main Results}
\textbf{State-of-the-art ASD models fall short on \datasetname{}.}  
Table~\ref{tab:models_on_our} shows that across a range of architectures, active speaker detection models trained on \datasetname{} achieve significantly lower mAP compared to their performance on AVA~\cite{roth2020ava}. For example, LoCoNet~\cite{wang2024loconet}, LASER~\cite{nguyen2025laser}, and TalkNCE~\cite{ts_talknet}, which report mAP scores above 95 on AVA, obtain only 82.2, 82.2, and 83.2 mAP, respectively, on \datasetname{}. Earlier models such as ASC~\cite{alcazar2020active} and ASDNet~\cite{kopuklu2021design} perform even worse, highlighting that \datasetname{} presents a substantially more challenging evaluation setting. These results suggest that state-of-the-art models, while successful on existing benchmarks, still fall short on \datasetname{}. This gap indicates that the ASD task remains unsolved under more realistic conditions, and that prior benchmarks may not fully capture the challenges faced in real-world deployments.

\textbf{State-of-the-art ASD models struggle across all axes of real-world difficulty.}  
As shown in Table~\ref{tab:models_on_our}, no model excels across any of the defined evaluation axes in \datasetname{}. Even the best-performing method, TalkNCE, achieves only moderate scores when faced with underrepresented languages (86.7), noisy backgrounds (84.1), and crowded scenes (84.9), all lower than its nearly perfect mAP on AVA~\cite{roth2020ava}, indicating that these real-world conditions remain challenging individually. The performance drops even further in the Hard subset (77.9), where multiple challenges co-occur. We also benchmark a TalkNCE model trained on AVA~\cite{roth2020ava}, which performs poorly across the board on \datasetname{}, scoring 77.5 overall and as low as 64.8 on Hard. These results highlight that existing ASD models are not only far from saturated under realistic settings, but that models trained on AVA generalize poorly when exposed to real-world acoustic and visual diversity.

\begin{table*}[t]
  \centering
  \footnotesize
  \caption{\textbf{Generalization of models trained on \datasetname{}.} We report mAP scores for each model evaluated on AVA, Talkies, and ASW after training on \datasetname{}. Results on \datasetname{} represent in-domain performance, while the others reflect generalization to out-of-domain benchmarks. The consistently strong performance across all benchmarks indicates that \datasetname{} provides transferable learning signals that support robust ASD model development.}
  
  \begin{tabular}{cccccc}
    \toprule
    Model & Architecture & In-domain & \multicolumn{3}{c}{Out-of-domain} \\
    \cmidrule(r){4-6}
    & & \datasetname{} & AVA~\cite{roth2020ava} & Talkies~\cite{alcazar2021maas} & ASW~\cite{kim2021look} \\
    \midrule
    TalkNet~\cite{tao2021someone}         & ResNet/LIM        & 75.7 & 78.4 & 89.2 & 88.9 \\
    LoCoNet~\cite{wang2024loconet}        & TalkNet/SIM       & 82.2 & 84.4 & 91.0 & 90.0 \\
    LASER~\cite{nguyen2025laser}        & TalkNet/Lip Landmark       & 82.2 & 84.5 & 91.3 & 90.5 \\
    TalkNCE~\cite{jung2024talknce}        & LoCoNet/NCE loss  & 83.2 & 88.0 & 91.4 & 90.7 \\
    \bottomrule
  \end{tabular}

  \label{tab:models_on_all}
\end{table*}

\begin{figure*}[t]
  \centering
    \includegraphics[width=0.95\linewidth]{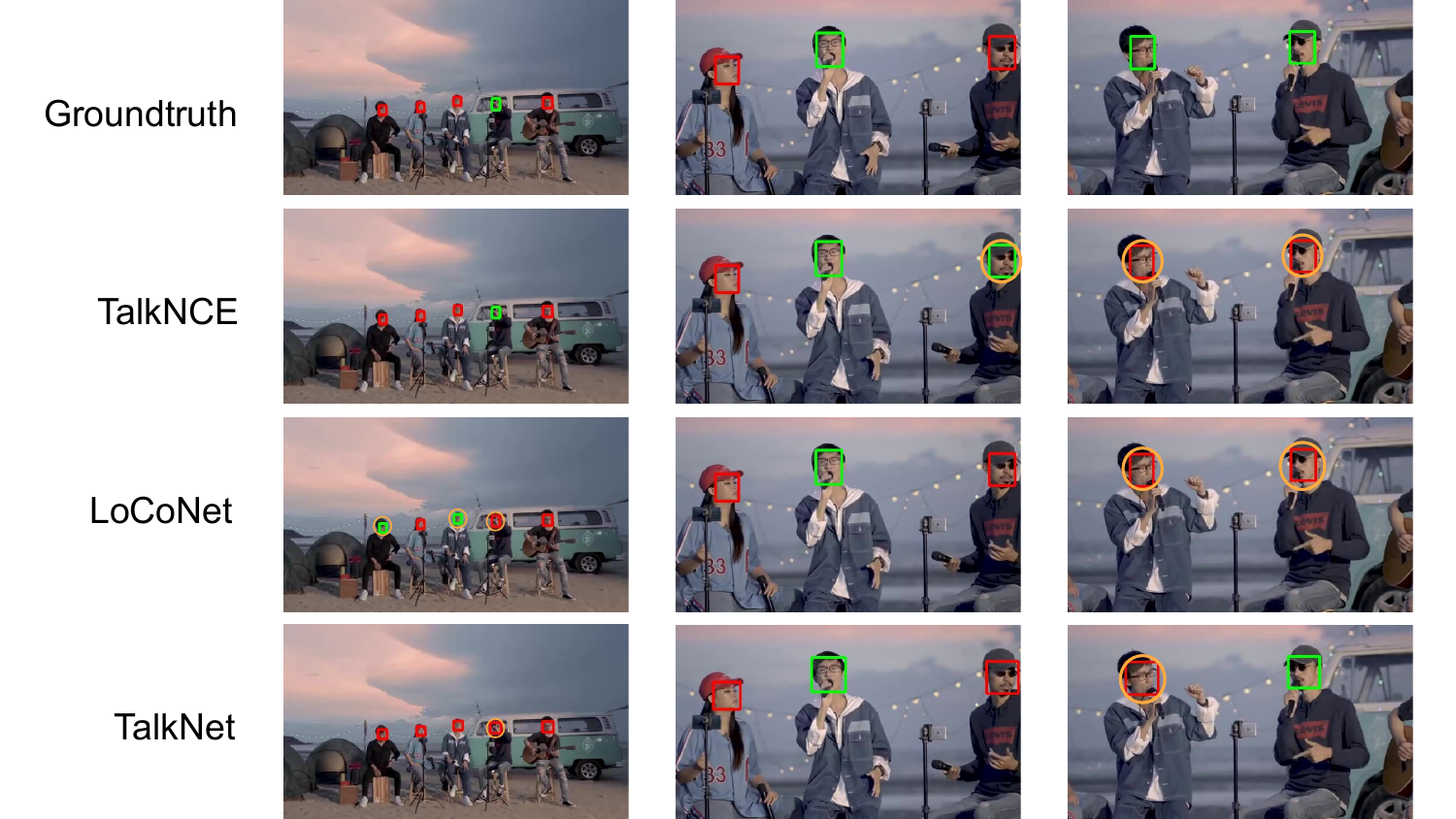}
    \caption{\textbf{Failure cases of state-of-the-art ASD models on a challenging test video from \datasetname{}.} This example contains all three difficulty axes: (1) \textit{crowded scenes} with multiple small and overlapping faces, (2) \textit{high background noise} from musical instruments and ambient sounds in an open environment, and (3) \textit{language variation}, with all speech in Vietnamese. All three models, TalkNCE, LoCoNet, and TalkNet, are not robust in this setting, with predictions frequently incorrect across frames. \textcolor{green}{Green} and \textcolor{red}{Red} indicate ground truth speaking and non-speaking frames; \textcolor{orange}{Orange} marks incorrect predictions.}

  \label{fig:fail}
\end{figure*}

\textbf{Models trained on \datasetname{} show strong cross-dataset transfer.}  
To evaluate whether the difficulty of \datasetname{} stems from noisy or unrealistic data, we evaluated the top four models from Table~\ref{tab:models_on_our} (TalkNet, LoCoNet, LASER, and TalkNCE) on three established ASD benchmarks: AVA~\cite{roth2020ava}, Talkies~\cite{alcazar2021maas}, and ASW~\cite{kim2021look}. As shown in Table~\ref{tab:models_on_all}, these models consistently achieve strong results across datasets despite never being trained on them. This suggests that \datasetname{} provides diverse and transferable learning signals, and that its increased difficulty reflects realistic variation rather than annotation noise or domain-specific artifacts.

\subsection{Comparing Benchmarks by Cross-Dataset Generalization Performance}

\textbf{Setup.}  
To compare the generalization characteristics of existing ASD benchmarks, we conduct a cross-dataset experiment using the state-of-the-art ASD framework: LoCoNet~\cite{wang2024loconet} + TalkNCE loss~\cite{jung2024talknce}. We train the model independently on each of the four datasets AVA~\cite{roth2020ava}, Talkies~\cite{alcazar2021maas}, ASW~\cite{kim2021look}, and \datasetname{}, and we evaluate the performance on all four benchmarks. The results are summarized in Table~\ref{tab:cross}.

\textbf{Results.}  
We observe a striking contrast in generalization behavior. Models trained on existing datasets like AVA, Talkies, and ASW achieve near-perfect in-domain performance (e.g., over 95 mAP), but exhibit substantial performance drops when evaluated on any other dataset. This indicates that such models tend to overfit to dataset-specific cues, which may be unrepresentative of broader real-world variability. In contrast, the best performing model, TalkNCE, trained on \datasetname{} does not achieve saturated performance in-domain (83.2 mAP), but generalizes significantly better to the other three datasets, with strong mAP scores of 88.0, 91.4, and 90.4 on AVA, Talkies, and ASW, respectively. These results suggest that prior benchmarks cover a limited range of speaking scenarios, enabling models to overfit to narrow acoustic and visual patterns. \datasetname{}, by contrast, introduces richer scenario diversity, which encourages models to learn more robust and transferable representations. This makes \datasetname{} not only a more challenging benchmark, but also a more effective training source for models intended for deployment in realistic, unconstrained environments.

\begin{table}[t]  
  \centering
  \footnotesize
  \caption{\textbf{Cross-dataset generalization comparison.} Each row reports mAP for a TalkNCE model trained on the specified dataset and evaluated on all four benchmarks. Prior datasets show strong in-domain but weak cross-dataset performance, whereas \datasetname{} enables consistent generalization.}
  
  \begin{tabular}{ccccc}
    \toprule        
    Train\textbackslash Eval & AVA~\cite{roth2020ava} & Talkies~\cite{alcazar2021maas} & ASW~\cite{kim2021look} & \datasetname{} \\
    \midrule
    AVA~\cite{roth2020ava}       & 95.5 & 88.3 & 88.5 & 77.5 \\
    Talkies~\cite{alcazar2021maas} & 55.7 & 95.6 & 84.5 & 59.9 \\
    ASW~\cite{kim2021look}       & 29.2 & 58.8 & 96.1 & 33.8 \\
    \datasetname{}               & 88.0 & 91.4 & 90.4 & 83.2 \\
    \bottomrule
  \end{tabular}
  \vspace{-3mm}
  \label{tab:cross}
\end{table}

\begin{figure}[h]
  \centering
    \includegraphics[width=\columnwidth]{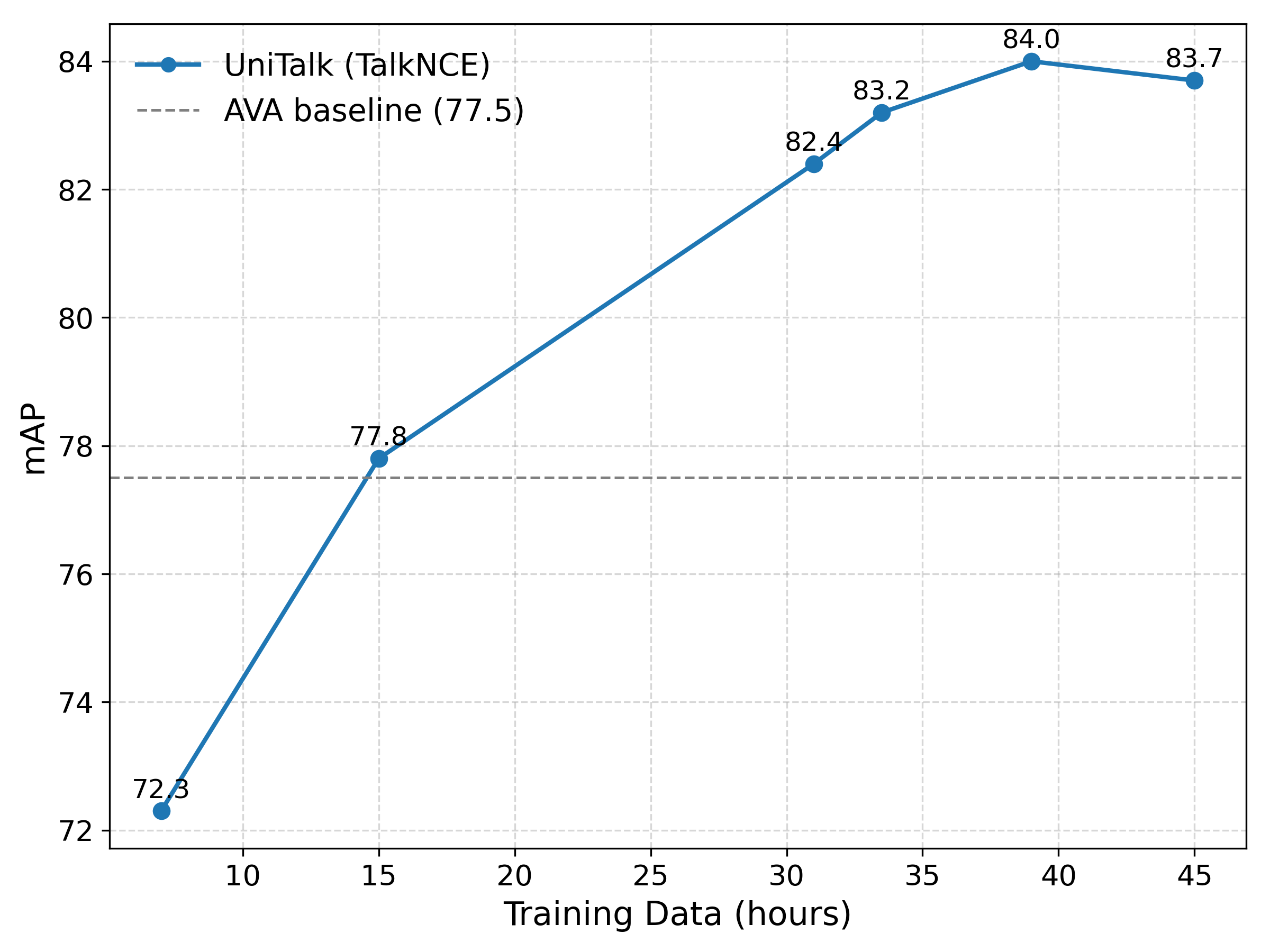}
  \caption{Data scaling study to evaluate tradeoff between performance and efficiency using TalkNCE.  Model performance increases up to around 33.5 hours of training data, after which additional data yields only marginal gains.}
  \label{fig:data_scale}
\end{figure}

\subsection{\datasetname{} as a Pretraining Source}

\textbf{Setup.}
To assess the utility of \datasetname{} as a pretraining source, we examine how effectively a model trained on \datasetname{} can adapt to AVA~\cite{roth2020ava} using limited additional data. Specifically, we fine-tune the TalkNCE model~\cite{jung2024talknce} - initially trained on \datasetname{} - on varying amounts of AVA training data, ranging from 3 to 15 hours of video, as well as the full AVA training set. For each setting, we report mAP on both AVA (the target domain) and \datasetname{} (the original domain) to evaluate adaptation and knowledge retention.

\textbf{Results.}
As shown in Table~\ref{tab:finetune}, the model rapidly adapts to AVA, achieving 92.4 mAP with only 3 hours of AVA training data. Performance continues to improve with additional data, reaching 95.7 mAP with the full dataset. Throughout this process, performance on \datasetname{} remains strong, showing no significant degradation. These results indicate that \datasetname{} serves not only as a challenging evaluation benchmark but also as an effective pretraining source. The model acquires transferable representations that can be quickly adapted to narrower domains such as AVA, making it a practical starting point for real-world ASD applications.

\begin{table}[t]
  \centering
  \caption{\textbf{Fine-tuning a TalkNCE model~\cite{jung2024talknce} pretrained on \datasetname{} using AVA~\cite{roth2020ava}.}
  Each row reports mAP after fine-tuning on a different amount of AVA training data (measured in video hours). The model quickly adapts to AVA while maintaining strong performance on \datasetname{}.}

  \label{tab:finetune}
  \resizebox{\columnwidth}{!}{%
  \begin{tabular}{@{}lcccc@{}}
    \toprule
    Pretraining Data & AVA Training Hours & Epochs & AVA~\cite{roth2020ava} & \datasetname{} \\
    \midrule
    None             & 31hr (full AVA)    & 25     & 95.5 & 77.5 \\
    \midrule
    \datasetname{}   & 3hr                 & 2      & 92.4 & 80.4 \\
    \datasetname{}   & 5hr                 & 5      & 93.4 & 78.6 \\
    \datasetname{}   & 10hr                & 10     & 94.0 & 79.2 \\
    \datasetname{}   & 15hr                & 15     & 95.0 & 80.6 \\
    \datasetname{}   & 31hr (full AVA)     & 15     & 95.7 & 81.3 \\
    \bottomrule
  \end{tabular}%
  \vspace{-1cm}
  }
\end{table}

\subsection{Data Scaling}

Finally, we conduct a controlled scaling study by annotating a total of 44.5 hours of training data. Mean average precision increases by 0.8 mAP from 31 hours to 33.5 hours and by another 0.8 mAP from 33.5 hours to 39 hours, then plateaus and slightly declines at 45 hours (Figure~\ref{fig:data_scale}). This saturation suggests diminishing returns from additional data under our setup. Balancing benchmark difficulty, realism, and computational accessibility, we therefore adopt 33.5 hours for training and 11 hours for evaluation. We also provide an additional 11 hours as a supplementary training set for future model development.

\subsection{Qualitative Failure Cases}

Figure~\ref{fig:fail} presents an example from the \datasetname{} test set where multiple state-of-the-art active speaker detection (ASD) models fail, despite being trained on \datasetname{} itself. The video combines all three major difficulty factors targeted by our benchmark: (1) an underrepresented language (Vietnamese), (2) high background noise-including musical elements and ambient context sounds from an open public space-and (3) a visually crowded scene with multiple visible faces.

This example highlights that even leading models struggle when multiple real-world challenges are present simultaneously. It underscores that the task is not saturated and validates the need for benchmarks like \datasetname{} that explicitly test robustness under diverse and overlapping sources of difficulty.

\section{Limitation and Future Work}

\datasetname{} improves language diversity over prior benchmarks such as AVA, but its distribution remains imbalanced due to practical curation constraints. English is overrepresented because public video platforms offer more high-quality, diverse English content. In many other languages, suitable data were harder to curate because of ethical/safety exclusions, keyword-based sourcing limits, and our limited fluency for validation. As a result, \datasetname{} does not achieve perfect linguistic balance. To partially address this, we provide a dedicated evaluation subgroup for underrepresented languages to measure model performance in less-represented settings. Future versions may improve coverage through multilingual collaboration and broader community contributions.

\section{Conclusion}

We introduced \datasetname{}, a new large-scale benchmark for active speaker detection designed to better reflect the complexity of real-world audiovisual environments. Unlike prior benchmarks that rely heavily on clean, scripted movie content, \datasetname{} emphasizes diversity across three key axes of difficulty: language variation, background noise, and visual crowding, offering a more realistic and challenging testbed for model development. Through extensive experiments, we show that while state-of-the-art models achieve near-perfect scores on existing datasets like AVA, their performance drops significantly on \datasetname{}. Moreover, our results highlight that models trained on \datasetname{} generalize more robustly across other real-world datasets, demonstrating the value of \datasetname{} as both a benchmark and a pretraining resource. Subgroup analyses further reveal persistent model weaknesses under specific difficulty conditions, motivating targeted improvements. We hope \datasetname{} will serve as a valuable benchmark for the community and foster the development of more robust, generalizable active speaker detection models in realistic settings.

\newpage
\section{Generative AI Use Disclosure}
We used generative AI tools in a limited, assistive role during dataset construction and manuscript preparation. Specifically, AI was used to help brainstorm and expand seed keywords/search phrases that were then used to instantiate keyword-based searches for candidate video categories (e.g., scene types likely to contain targeted audiovisual conditions). All generated keywords were manually reviewed, filtered, and refined by the authors before use in data sourcing.

We also used AI-assisted writing tools to polish the manuscript’s grammar and vocabulary for clarity and readability. AI tools were not used to generate annotations, experimental results, model outputs, scientific claims, or final conclusions. All methodological decisions, data curation judgments, analyses, and interpretations were made and verified by the authors, who take full responsibility for the content of this paper.

\section{Acknowledgement}
This work was supported in part by NSF IIS2404180, IBM, and Institute of Information \& communications Technology Planning\& Evaluation (IITP) grant funded by the Korea government (MSIT) ((No. 2022-0-00871, Development of AI Autonomy and Knowledge Enhancement for AI Agent Collaboration), (No.RS-2025-02219317, AI Star Fellowship (Kookmin University)).
\bibliographystyle{IEEEtran}
\bibliography{mybib}

\end{document}